\documentclass[10pt,twocolumn,letterpaper]{article}

\usepackage[pagenumbers]{cvpr} % To force page numbers, e.g. for an arXiv version

\usepackage{xcolor}
\usepackage{multirow}
\usepackage{makecell}

\definecolor{cvprblue}{rgb}{0.21,0.49,0.74}
\usepackage[pagebackref,breaklinks,colorlinks,citecolor=cvprblue]{hyperref}

\def\confName{CVPR}
\def\confYear{2025}
\def\modelname{MambaIC}

\title{\modelname{}: State Space Models for High-Performance Learned \\ Image Compression}

\author{Fanhu Zeng$^{1}$\thanks{Work done during internship at AIR, Tsinghua University}, 
	Hao Tang$^{2}$,
    Yihua Shao$^{3}$,
    Siyu Chen$^{3}$,
    Ling Shao$^{4}$,
    Yan Wang$^{1}$\thanks{Corresponding author. \small $<$wangyan@air.tsinghua.edu.cn$>$}\\
	$^{1}$Institute for AI Industry Research~(AIR), Tsinghua University\\
    $^{2}$School of Computer Science, Peking University \\
    $^{3}$University of Science and Technology Beijing\\
    $^{4}$UCAS-Terminus AI Lab, University of Chinese Academy of Sciences \\
}

\begin{document}
\maketitle

\begin{abstract}
A high-performance image compression algorithm is crucial for real-time information transmission across numerous fields. Despite rapid progress in image compression, computational inefficiency and poor redundancy modeling still pose significant bottlenecks, limiting practical applications. Inspired by the effectiveness of state space models~(SSMs) in capturing long-range dependencies, we leverage SSMs to address computational inefficiency in existing methods and improve image compression from multiple perspectives. In this paper, we integrate the advantages of SSMs for better efficiency-performance trade-off and propose an enhanced image compression approach through refined context modeling, which we term \textbf{\modelname{}}. Specifically, we explore context modeling to adaptively refine the representation of hidden states. Additionally, we introduce window-based local attention into channel-spatial entropy modeling to reduce potential spatial redundancy during compression, thereby increasing efficiency. Comprehensive qualitative and quantitative results validate the effectiveness and efficiency of our approach, particularly for high-resolution image compression. Code is released at \url{https://github.com/AuroraZengfh/MambaIC}.
\end{abstract}

\section{Introduction}
\label{sec:intro}
Image compression is a crucial aspect of image storage and transmission, particularly in the era of high-definition, high-resolution, and large-scale digital images. In recent decades, traditional image formats such as JPEG~\cite{pennebaker1992jpeg}, BPG~\cite{bellard2015bpg}, and VVC~\cite{bross2021overview} have significantly advanced the digitalization process. With the rapid development of modern architectures like CNNs~\cite{he2016deep, tan2019efficientnet}, Transformers~\cite{dosovitskiy2020image, kenton2019bert}, and others~\cite{sun2024learning, peng2023rwkv}, many learned image compression~(LIC) methods have emerged to achieve a better performance-efficiency trade-off. These methods typically consist of an encoder, decoder, and entropy model, and are optimized end-to-end. Recent works~\cite{liu2023learned,he2022elic,jiang2023mlicpp} even surpass the classical standards~\cite{bross2021overview}, demonstrating their potential to lead next-generation image compression standards and drive real-world applications in related industries.

\begin{figure}[t]
    \centering
    \includegraphics[width=0.95\linewidth]{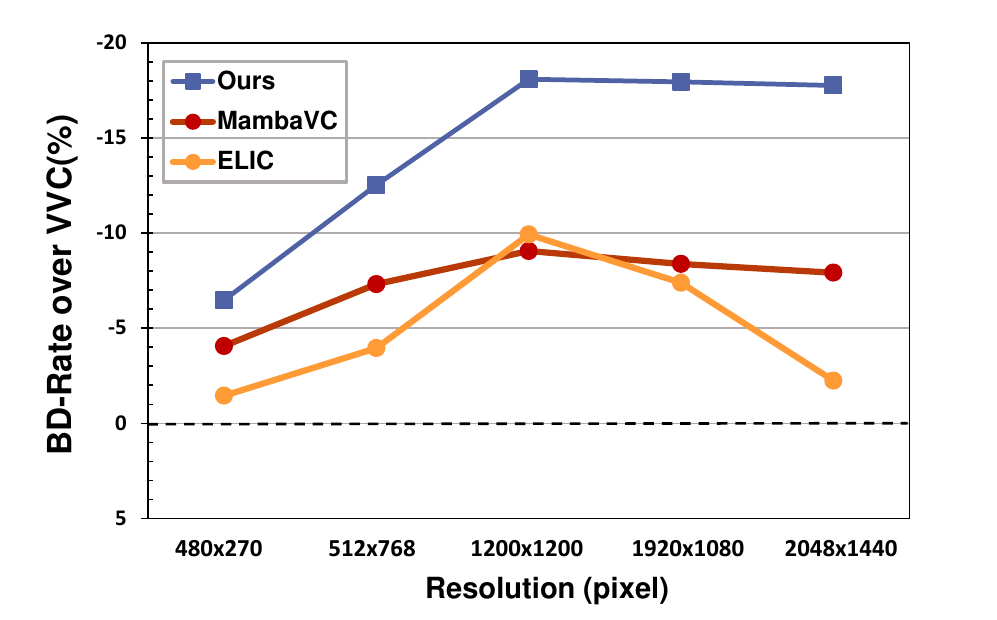}
    \vspace{-5pt}
    \caption{Performance of typical image compression approaches when resolution varies. Notably, our \modelname{} consistently outperforms existing state-of-the-art methods~\cite{he2022elic,qin2024mambavc} and gets a more substantial improvement when the image scales up.}
    \vspace{-10pt}
    \label{fig:efficiency-performance-illustration}
\end{figure}

Despite the rapid advancements achieved by pioneering works in both CNNs~\cite{balle2018variational,he2022elic} and Transformers~\cite{qian2022entroformer, koyuncu2022contextformer}, limitations in performance and computational complexity hinder their practical applications. Specifically, while Transformer-based methods generally outperform CNN-based methods, they suffer from quadratic complexity proportional to pixel numbers, resulting in significant latency particularly for high-resolution image compression. Although efforts are made~\cite{liu2023learned,jiang2023mlicpp} with well-designed modules to reduce computational load, the balance between performance and efficiency remains unsolved. This highlights the need for improving efficiency while maintaining high-performance in image compression.

In contrast to designing delicated modules, another way to enhance LIC is to employ novel architectures. For example, introducing Transformer into LIC~\cite{qian2022entroformer} brings significant improvements. Recent advances like State Space Models~(SSMs) have demonstrated significant effectiveness, partly due to their reduced complexity and hardware-friendly computation. Therefore, various SSM variants have been applied to numerous downstream tasks~\cite{hatamizadeh2024mambavision, zhu2024vim, yang2024vivim}. As a novel paradigm, it offers inherent advantages over CNNs and transformers and paves a novel way for LIC. In terms of exploiting recent architectures for image compression, one concurrent work is MambaVC~\cite{qin2024mambavc}, which attempts to improve the efficiency of compression operations with an SSM structure. However, it merely replaces foundational blocks with SSM blocks without adjustments attached to the specific characteristics of SSM, resulting in sub-optimal performance compared to other well-designed image compression architectures.

In this work, we address two key challenges in image compression: \textbf{\textit{efficiency and performance}}. To achieve these goals, we utilize SSMs to improve compression performance while enhancing computational efficiency, which can be validated by performance improvement of different resolutions illustrated in~\Cref{fig:efficiency-performance-illustration}. We employ SSMs into image compression framework and promote high-performance learned image compression from several perspectives. Specifically, we integrate SSMs into the context model to enrich the side information, thereby improving the efficiency and effectiveness of encoding and decoding. Additionally, we use channel-spatial entropy modeling with window-based local attention to better capture redundancy in latent representations, further boosting performance. We conduct extensive experiments, and the results demonstrate that the proposed method is both effective and efficient. Notably, \modelname{} shows excellent efficiency in high-resolution image compression. Comprehensive analysis and visualization further certify its usefulness.

Our contributions can be summarized as follows:
\begin{itemize}
    \item To the best of our knowledge, we are the first to integrate SSMs into both nonlinear transform and context model for high-performance learned image compression.
    \item We incorporate window-based local attention into channel-spatial entropy modeling to reduce spatial redundancy as well as enhancing the compression pipeline.
     \item We conduct comprehensive experiments and analyses to demonstrate that the proposed method is effective, efficient with superior qualitative and quantitative results, particularly for high-resolution images.
\end{itemize}

\section{Related Work}
\subsection{State Space Models}
State Space Models~(SSMs)~\cite{mehta2023long, gu2021combining} is becoming an emerging and promising direction for modern architecture applications. It aims to tackle the problem of computational efficiency in modeling long-range dependencies~\cite{vaswani2017attention, dosovitskiy2021image} from a novel perspective and seeks to process long sequences with linear-time complexity. Following the pioneering structured state-space model~(S4)~\cite{gu2022efficiently}, numerous works~\cite{fu2022hungry,smith2022simplified} improve the performance.

As a notable milestone, Mamba~\cite{gu2023mamba, dao2024transformers} introduces an input-dependent selection technique into S4 and largely enhances SSM with a novel structure. Due to its impressive potential, many efforts have been paid to promote the scanning mechanism to refine long-range modeling in vision tasks~\cite{liu2024vmamba, zhu2024vim, hatamizadeh2024mambavision, huang2024localmamba}. Specifically, Vim~\cite{liu2024vmamba} incorporates bidirectional sequence modeling with positional awareness for visual understanding. Mambavision~\cite{hatamizadeh2024mambavision} proposes a redesigned, vision-friendly, hybrid Mamba block to enhance global context learning. LocalMamba~\cite{huang2024localmamba} introduces a local scan with distinct windows to enhance visual representations. 
The success of SSMs has also inspired widespread application in various downstream tasks of the vision domain~\cite{guan2024qmamba,yang2024vivim,fei2024dimba,shi2024vmambair}. In this work, our aim is to specifically design the structure of SSMs to leverage the power for image compression.

\subsection{Learned Image Compression}
Compression~\cite{zeng2024m2m, wu2023ppt, shao2025trdq} plays a vital role in acceleration and learned image compression~\cite{minnen2018joint} targets optimizing rate-distortion trade-off exploiting neural networks as image compression operator. Modern learned image compression framework fundamentally constructs upon entropy model~\cite{balle2017end}, followed by employing hyperprior~\cite{balle2018variational} to calculate the entropy parameters and many specific structural improvements~\cite{chen2022two, cheng2020learned, zou2022devil}.

With the widespread adoption of Transformer~\cite{vaswani2017attention,dosovitskiy2021image}, they are also used in image compression~\cite{lu2022transformer}. However, computational efficiency remains a critical problem due to the quadratic growth time complexity despite the remarkable progress in performance and many works seek to tackle the problem with fine architecture design~\cite{he2022elic,liu2023learned,jiang2023mlicpp}. As a concurrent exploration, MambaVC~\cite{qin2024mambavc} first attempts to build a compression framework based on SSM~\cite{gu2023mamba}. However, it employs mamba blocks without special adaptation, leading to inferior performance. In this paper, we comprehensively analyze the characteristics of SSMs in compression and enhance image compression with better rate-distortion performance.

\subsection{Context Models}

\begin{figure*}[t]
    \centering
    \includegraphics[width=0.95\linewidth]{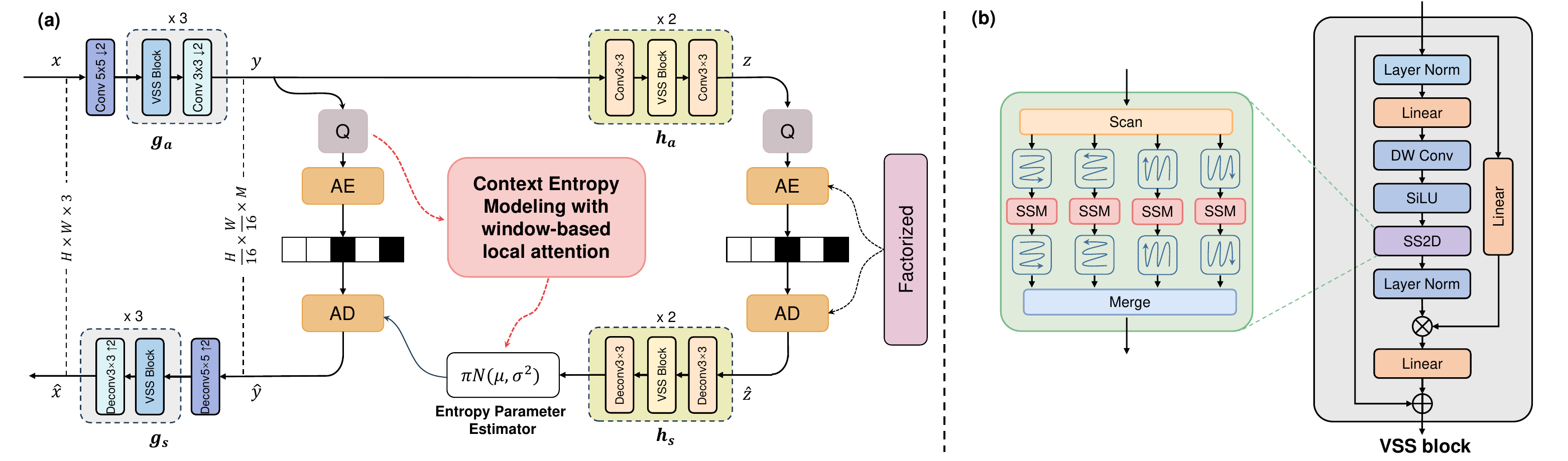}
    \caption{(a) Overall architecture of the proposed method. AE and AD are arithmetic encoder/decoder with residual bottleneck blocks. The proposed context entropy model consists of channel-spatial context model and window-based local attention to estimate entropy parameters. (b) Structure of visual state space~(VSS) block with 2D selective scan.} 
    \label{fig:overall-structure}
    \vspace{-15pt}
\end{figure*}

Many works have focused on reducing bitstream with context model. \citet{minnen2018joint} proposes to estimate current representations using previously coded ones for better compression performance. Moreover, with impressive progress accomplished by channel-wise and checkerboard auto-regressive models~\cite{minnen2020channel, he2021checkerboard, he2022elic} that split the representation into groups and encode each group conditioned on coded symbols of previous groups, some progress with respect to Transformer~\cite{qian2022entroformer,koyuncu2022contextformer,zhu2022transformer} also gains promotion. 

Despite the achievements, computational complexity still remains a challenge for existing approaches. For the first time, we explore an efficient context model that incorporates SSMs for learned image compression to achieve competitive results against existing CNN/transformer methods with better computational efficiency. 

\section{The Proposed Method}
\label{sec:method}

We begin with problem formulation and notation for clear statement in~\Cref{sec:preliminary}. Then we describe the proposed method termed \modelname{} in~\Cref{sec:mambaic}, which is composed of SSM-based nonlinear transform, context model, and window-based local attention for channel-spatial entropy modeling in an attempt to improve computational efficiency and enhance effectiveness. The overall pipeline and detailed module structure are shown in~\Cref{fig:overall-structure}.

\subsection{Problem Formulation and Notations}
\label{sec:preliminary}
\noindent \textbf{Learned image compression.} 
Typically, a given input image $\boldsymbol{x}$ is first encoded into latent representation $\boldsymbol{y}$ through neural analyzer $g_a$ parametered by $\boldsymbol{\theta}_{g_a}$:
\begin{equation}
    \boldsymbol{y} = g_a(\boldsymbol{x}; \boldsymbol{\theta}_{g_a}).
\end{equation}

Hyper prior encoder/decoder $h_a$ and $h_s$ are employed to learn the mean and variance of latent features by hyperprior representation $\boldsymbol{z}$ extracted from $\boldsymbol{y}$:
\begin{equation}
\begin{aligned}
    \boldsymbol{z} &= h_a(\boldsymbol{y}; \boldsymbol{\theta}_{h_a}),\\
    \boldsymbol{\mu},  \boldsymbol{\sigma^2} &= h_s(\boldsymbol{\hat{\boldsymbol{z}}},\boldsymbol{\theta}_{h_s}).
\end{aligned}
\end{equation}

The discrete coding-symbols $\hat{\boldsymbol{y}}$ and $\hat{\boldsymbol{z}}$ are obtained from quantized latent and hyperprior representations:
\begin{equation}
\begin{aligned}
    \hat{\boldsymbol{y}} &= \mathcal{Q}(\boldsymbol{y}-\boldsymbol{\mu}) +\boldsymbol{\mu}, \\
    \hat{\boldsymbol{z}} & = \mathcal{Q}(\boldsymbol{z}),
\end{aligned}
\end{equation}
where $\mathcal{Q}(\cdot)$ stands for quantization operator, and entropy model is formulated as conditioned gaussian form to estimate the conditional distribution of latent representation:
\begin{equation}
    p_{\hat{\boldsymbol{y}}|\hat{\boldsymbol{z}}}(\hat{\boldsymbol{y}}|\hat{\boldsymbol{z}}) = \left [ \mathcal{N}(\boldsymbol{\mu},\boldsymbol{\sigma^2}) * U(-0.5,0.5)) \right ](\hat{\boldsymbol{y}}),
\end{equation}
and $\hat{\boldsymbol{x}}$ is the reconstructed images in pixel space from neural synthesizer $g_s$ parametered by $\boldsymbol{\theta}_{g_s}$:
\begin{equation}
    \hat{\boldsymbol{x}} = g_s(\hat{\boldsymbol{y}},\boldsymbol{\theta}_{g_s}).
\end{equation}

During training, lagrangian multiplier $\lambda$ is introduced to adjust the rate-distortion optimization and control bit rate in an end-to-end manner:
\begin{equation}
    \mathcal{L} = \lambda \mathcal{D}(\boldsymbol{x}, \hat{\boldsymbol{x}})+\mathcal{R}(\hat{\boldsymbol{y}}) + \mathcal{R}(\hat{\boldsymbol{z}}),
\end{equation}
where $\mathcal{D}(\boldsymbol{x}, \hat{\boldsymbol{x}})$ measures the distortion by mean squared error~(MSE) and $\mathcal{R}(\hat{\boldsymbol{y}})$, $\mathcal{R}(\hat{\boldsymbol{z}})$ are bit rates of $\hat{\boldsymbol{y}}$ and $\hat{\boldsymbol{z}}$ estimated by entropy model.

\noindent \textbf{State space models.}
SSMs~\cite{gu2022efficiently} can be viewed as linear time-invariant~(LTI) systems. Dynamic response $\boldsymbol{y}(t)\in\mathbb{R}^L$ is stimulated from input $\boldsymbol{x}(t)\in\mathbb{R}^L$ through intermediate hidden state $\boldsymbol{h}(t)\in\mathbb{R}^N$, which can be described in the form of linear ordinary differential equations~(ODEs):
\begin{equation}
 \begin{aligned}
    \boldsymbol{h}'(t) &=  \mathbf{A} \boldsymbol{h}(t) + \mathbf{B}\boldsymbol{x}(t),\\
     \boldsymbol{y}(t) &=  \mathbf{C} \boldsymbol{h}(t),
 \end{aligned}
 \label{eqn:continuous-ssm}
\end{equation}
where $\mathbf{A}\in\mathbb{R}^{N\times N}$, $\mathbf{B}\in\mathbb{R}^{N\times 1}$, $\mathbf{C}\in\mathbb{R}^{N\times 1}$ are state transition matrices, respectively. Under the zero-order hold assumption that the value of input $\boldsymbol{x}$ is constant over the interval $\Delta$, The parameters of continuous system can be discretized into the following form:
\begin{equation}
\begin{aligned}
    \bar{\mathbf{A}} & = \mathrm{exp}(\Delta\mathbf{A}), \\ 
    \bar{\mathbf{B}} & = (\Delta\mathbf{A})^{-1}(\mathrm{exp}(\Delta\mathbf{A})-\mathbf{I})\cdot \Delta\mathbf{B}.
\end{aligned}
\end{equation}

\begin{figure*}[ht]
    \centering
    \includegraphics[width=1.0\linewidth]{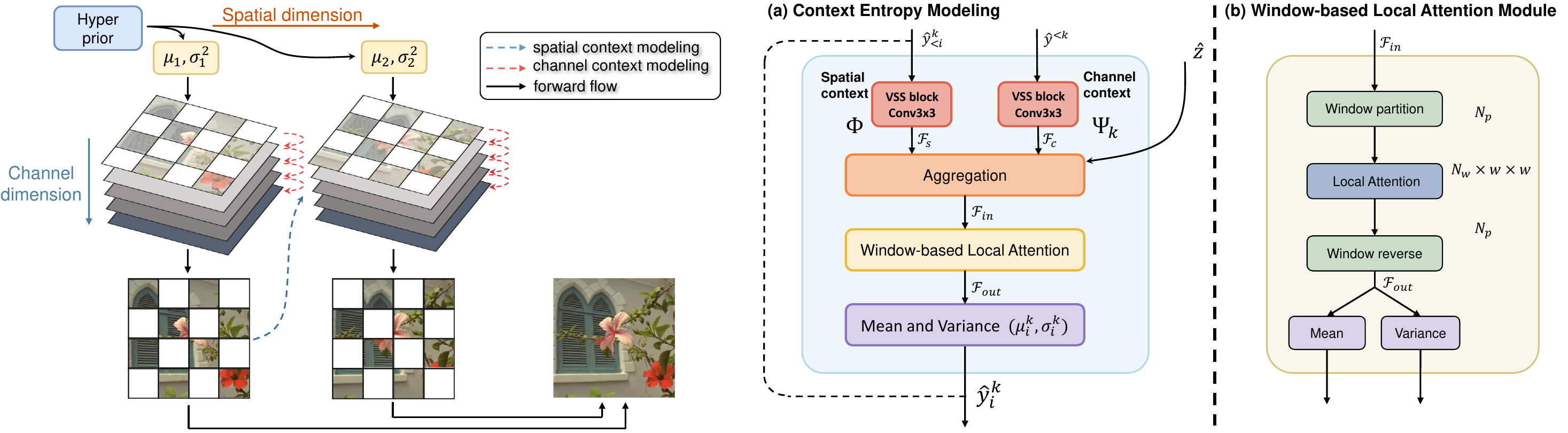}
    \caption{\textit{Left:} Illustration of hybrid context modeling. Hyperprior together with symbols of previous channels estimates the parameters of anchors, which are used by the prediction of non-anchors spatially. Both anchors and non-anchors are integrated to generate the entire latent representations. \textit{Right:} Detailed procedure of (a) auto-regressive context entropy modeling with (b) window-based local attention.}
    \label{fig:context}
    \vspace{-15pt}
\end{figure*}

Consequently, the discrete equation of \cref{eqn:continuous-ssm} can be reformulated as:
\begin{equation}
    \begin{aligned}
        \boldsymbol{h}_t & =  \bar{\mathbf{A}}\boldsymbol{h}_{t-1} + \bar{\mathbf{B}}\boldsymbol{x}_t, \\
        \boldsymbol{y}_t &=  \mathbf{C}\boldsymbol{h}_t.
    \end{aligned}
\end{equation}

Subsequently, a global convolution is incorporated for parallel computation.

\noindent \textbf{Context model with hybrid auto-regression.} To enhance the entropy model and mitigate redundancy, context learning~\cite{minnen2018joint} separates the spatial information and recovers the features in an auto-regressive convolution. For latent representations $\hat{\boldsymbol{y}}_i$, context model refers to neighbours with spatial relation:
\begin{equation}
    \hat{\boldsymbol{y}}_i = \Phi(\hat{\boldsymbol{y}}_{<i};\boldsymbol{\theta}),
\end{equation}
where $\Phi$ is used to estimate the entropy parameters with hyperprior $\hat{\boldsymbol{z}}$. As an improvement, checkerboard~\cite{he2021checkerboard} groups them into anchors and non-anchors and therefore replace the auto-regressive decoding into masked latent modeling, which significantly accelerates the process.

In addition to spatial context modeling, channel-wise auto-regressive modeling~\cite{minnen2020channel} further splits the channels into $K$ chunks, and predicts the symbols on context of previous channels to enhance dimensional expression:
\begin{equation}
    \hat{\boldsymbol{y}}^k = \Psi_k(\hat{\boldsymbol{y}}^{<k},\boldsymbol{\theta_k}), \quad k =2,\cdots,K,
\end{equation}
where $\Psi_k$ is entropy parameter estimator of the $k^{th}$ chunk along channel dimension. 

Built upon this, ELIC~\cite{he2022elic} further integrates both checkerboard and channel-wise autoregressive modeling, and the resultant hybrid context model is illustrated in left of ~\Cref{fig:context}. In this paper, we also combine both advantages and exploit channel-spatial context model with uniform chunks along dimension.

\subsection{Overall Architecture of \modelname{}}
\label{sec:mambaic}
Due to the efficiency of SSM shown in previous research~\cite{gu2023mamba,zhu2024vim}, we employ it as the foundation component and explore better mechanisms tailored for the adaptation in image compression. Generally, we employ 2D selective scan as the foundation block in nonlinear transform and context model to predict the mean $\boldsymbol{\mu}$ and the variance $\boldsymbol{\sigma}$ of the Gaussian entropy model~\cite{minnen2018joint} and decode latent representations with fine window-based local attention through arithmetic decoder~(AD). The overall structure of the proposed method is shown in~\Cref{fig:overall-structure}\textcolor{red}{a}.

\noindent \textbf{SSM-based contextual model.} For SSM block, we utilize visual state space~(VSS)~\cite{zhu2024vim, hatamizadeh2024mambavision} block in parameter estimator for context modeling. As shown in~\Cref{fig:overall-structure}\textcolor{red}{b}, visual state space block consists of a composition of layer normalization, linear layer, depthwise convolution~(DW Conv), SiLU activation and the core 2D selective scan~(SS2D). Concretely, SS2D first scans the input patches along four distinct traversal paths, applies a separate SSM block to process each patch, and then merges them back to 2D output. Through cross-scanning and merging, SS2D effectively integrates information from all relevant pixels in 2D space and enhances global receptive fields, which greatly enhances image compression with advanced representation. 

\begin{figure*}[htbp]
    \centering
    \includegraphics[width=0.90\linewidth]{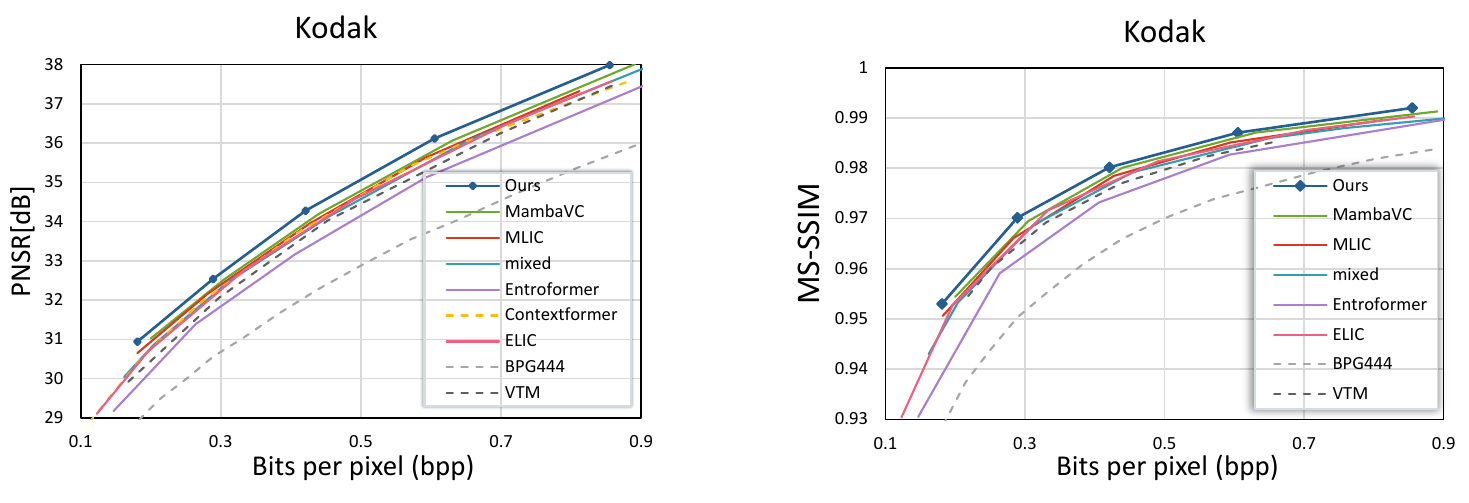}
    \vspace{-10pt}
    \caption{Results on Kodak. Performance is measured with PSNR and MS-SSIM. Left-top represents better performance. We compare with numerous learned compression methods as well as conventional coding like BPG and VTM. Zoom in for a better view.}
    \label{fig:main-results-kodak}
    \vspace{-15pt}
\end{figure*}

We integrate visual state space block in contextual entropy modeling $\Psi_k$ and $\Phi$ for compact and effective bitstream representation in the channel-spatial direction as shown in~\Cref{fig:context}\textcolor{red}{a}. Explicitly, for decoding of the symbol at $i^{th}$ location of the $k^{th}$ chunks, $\Psi_k$ and $\Phi$ are incorporated for modeling the channel context $\hat{\boldsymbol{y}}^{<k}$ and spatial context $\hat{\boldsymbol{y}}^k_{<i}$, respectively. $\Psi_k$ and $\Phi$ consist of convolution block and VSS block to enrich channel-spatial context information, which can be formulated as:
\begin{equation}
\begin{aligned}
    \mathcal{F}_c = &\Psi_k(\hat{\boldsymbol{y}}^{<k}) =Conv(VSS(\hat{\boldsymbol{y}}^{<k})),\\
    \mathcal{F}_s = &\Phi(\hat{\boldsymbol{y}}^k_{<i}) = Conv(VSS(\hat{\boldsymbol{y}}^k_{<i})), 
\end{aligned}
\end{equation}
where $\mathcal{F}_c$ and $\mathcal{F}_s$ represents channel and spatial features, respectively. Compact latent representations with rich information are therefore obtained with estimated parameters from parameter aggregation effectively. 

The $\hat{\boldsymbol{y}}^k$ is then used to calculate the features of the next decoded symbol through the spatial/channel context modeling in the circle. Finally, all chunks of $\hat{\boldsymbol{y}}^k$ are concatenated, and obtain the decoded latent representation~$\hat{\boldsymbol{y}}$.

In practical implementation, we use a parallel checkerboard mask model~\cite{he2021checkerboard} for spatial context modeling, which means the latent representations are grouped into anchors and non-anchors. The context prior to $\hat{\boldsymbol{y}}_{<i}^k$ is $\varnothing$ for \textit{anchors} and anchors for \textit{non-anchors}, respectively.

\noindent \textbf{Window-based local attention for auto-regressive modeling.}
Proper entropy modeling and latent representation are vital for effective information compression. As SSM block effectively captures global receptive fields, we additionally perform local attention in context entropy modeling. As illustrated in~\Cref{fig:context}\textcolor{red}{b}, we incorporate window-based local attention~(WLA) following latent parameter aggregation. Specifically, WLA consists of window partition $\mathcal{W}_p$ that splits patches $N_p$ into smaller window size $N_w$, local attention~($\mathcal{WA}$ operating on $N_w \times w \times w$) that efficiently models local relation within windows and window reverse $\mathcal{W}_r$ to recover the representation formulated as:
\begin{equation}
    \mathcal{F}_{out} = \mathcal{W}_r(\mathcal{LA}(\mathcal{W}_p(\mathcal{F}_{in})).
\end{equation}
The window-based local attention has two advantages: (1) improve efficiency with attention on fewer patches; (2) enhance local relation description for better redundancy mitigation since visual state space block builds upon getting global receptive fields. Therefore, we are capable of facilitating both effectiveness and efficiency in the SSM-based image compression paradigm. 
 
Formally, mean and variance for entropy modeling in each context model is formulated as:
\begin{equation}
    (\boldsymbol{\mu}^k_i, \boldsymbol{\sigma}^k_i) = \mathrm{WLA}(\mathrm{ParamAgg}(\mathcal{F}_c,\mathcal{F}_s,h_s(\hat{\boldsymbol{z}}))),
\end{equation}
where $\mathrm{WLA}(\cdot)$ is window-based local attention. The estimated entropy parameters are then separately utilized for Gaussian entropy modeling of latent representations.

\section{Experiments}
\label{sec:experiments}

\begin{figure*}[htbp]
    \centering
    \includegraphics[width=0.90\linewidth]{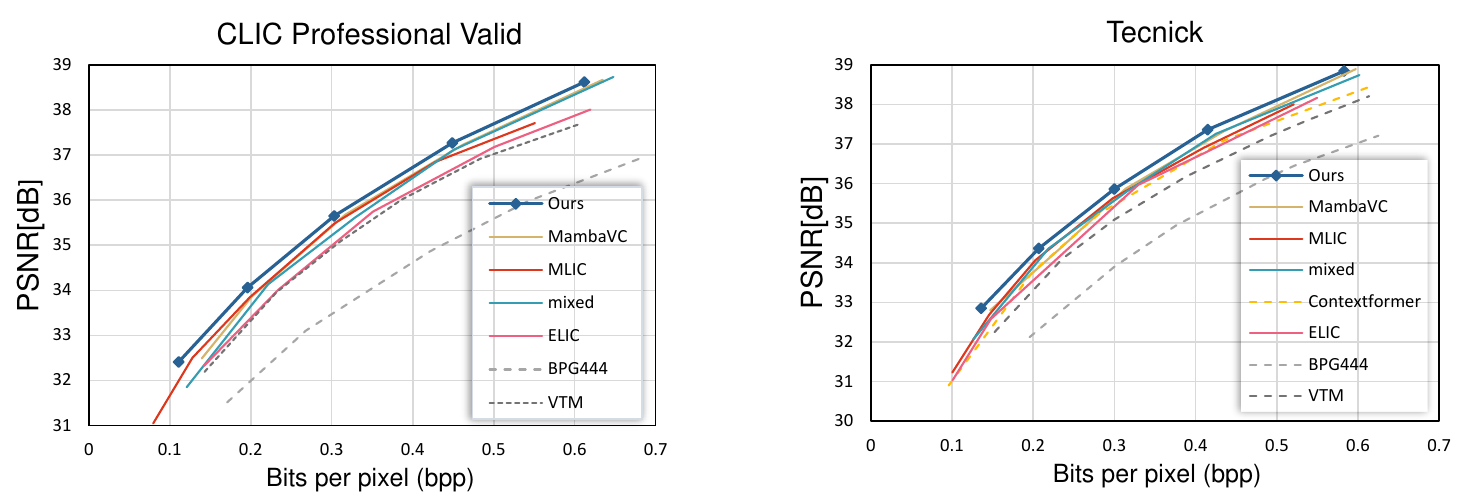}
    \vspace{-5pt}
    \caption{Performance on datasets with higher resolution including Tecnick and CLIC professional valid measured with PSNR. Left-top indicates better performance. Our method gets a consistent improvement when images scale up. Zoom in for a better view.}
    \label{fig:main-results-others}
    \vspace{-10pt}
\end{figure*}

\subsection{Setups}
\textbf{Baselines.} We undertake a comparison of our approach with multiple existing state-of-the-art image compression techniques for both CNN/Transformer-based methods including ELIC~\cite{he2022elic}, Mixed~\cite{liu2023learned}, MLIC~\cite{jiang2023mlicpp}, Entroformer~\cite{qian2022entroformer}, Contextformer~\cite{koyuncu2022contextformer} and MambaVC~\cite{qin2024mambavc}. Also, we include conventional coding methods BPG~\cite{bellard2015bpg} and VTM~\cite{bross2021overview} for a comprehensive comparison.

\noindent \textbf{Datasets.} For performance evaluation, we employ Kodak~\cite{franzen1999kodak}, Tecnick~\cite{asuni2014testimages} and CLIC Professional Valid~\cite{toderici2020workshop} as validation datasets for exhaustive evaluation. PSNR and MS-SSIM are used to comprehensively measure the distortion of the reconstructed images, and bits per pixel~(bpp) is used to measure compression ratio~(PSNR is expressed in logarithmic decibel units). For training, we use Flickr30k~\cite{flickr30k} which is composed of 31783 images. More details are shown in~\Cref{sec:appendix-metrics} and~\ref{sec:appendix-datasets}.

\subsection{Implementation Details}
The model is optimized with mean square error~(MSE) loss and we use a $\lambda \in \{0.0035,0.0067,0.013,0.025,0.05\}$ as the multiplier for different bitrates. We train the model for 250 epochs. The number of channels $N$ for $\boldsymbol{z}$ is 128. Number of channel $M$ for $\boldsymbol{y}$ is 320 and we set number of channel chunks $K$ to be 5. More implementation details are exhibited in~\Cref{sec:appendix-settings}.
\subsection{Rate-Distortion Performance}
\textbf{Comparison on common datasets.} We present the comparison with existing methods on commonly evaluated datasets in~\Cref{fig:main-results-kodak}. It is illustrated that our method achieves the best rate-distortion trade-off in both PSNR and MS-SSIM and consistently outperforms previous methods by a substantial margin, firmly verifying the effectiveness of our approach. Notably, compared to MambaVC with similar SSM structure, our method receives obvious improvements, showcasing that the proposed \modelname{} is capable of obtaining high performance against Transformer, CNN, and conventional coding algorithms while integrating the advantage of SSMs for image compression at the same time. 

\noindent \textbf{Scaling to higher resolution.} Due to the surge in demand for higher image sharpness, compression efficiency for images of higher resolution has received a large amount of attention. The primary advantage of SSMs is their effective computation, while maintaining global receptive fields. It is shown in \Cref{fig:main-results-others} that when the validation dataset varies to higher resolution~(1200$\times$1200 for Tecnick and 2048$\times$1440 for CLIC Professional Valid), our approach remains stable and gets competitive results against previous methods, strongly confirming the effectiveness and stability of the method. The conclusion is further confirmed in~\Cref{fig:efficiency-performance-illustration}, in which we present the BD-rate curve when the resolution of compressed images varies. It is shown that our method gets consistent improvements against existing methods. Specifically, the promotion becomes more substantial when the image scales up, where the performance of existing approaches shows varying degrees of degradation, whereas our method remains stable. It is of great significance in vast, high-definition, and large-scale image transmission in the era of big data.

\noindent \textbf{BD-rate comparison.} 
To have a quantitative comparison of the mentioned methods, we report BD-rate results anchored on the manually designed VVC image codec across different datasets. \Cref{tab:decoding} reveals that our approach outperforms conventional standard 
on RD performance regarding PSNR, and also surpasses existing methods by a substantial margin. Specifically, compared with typical context model in both CNN and transformer, our method gets impressive gains~(8.57\% and 7.47\%, respectively). Also, in comparison to SSM-based model MambaVC~\cite{qin2024mambavc}, considerable 5.21\% improvement strongly certificates the effectiveness of our method for high-performance image compression. 

\subsection{Efficiency Analysis}
To further substantiate the efficiency of \modelname, we compare it with various existing methods and systematically analyze the efficiency of the proposed method from the perspective of inference latency.

\begin{table}[t]
    \centering
    \caption{RD performance and complexity comparison of learned image compression evaluated on Kodak. Results marked with $\dag$ are from the original paper due to the absence of code. Typical CNN/Transformer methods are selected for comparative analysis. Arithmetic decoding time is included in decoding latency.}
    \vspace{-5pt}
    \renewcommand{\arraystretch}{1.25}
    \resizebox{\linewidth}{!}{
    \begin{tabular}{c cccc} 
    \toprule[1.3pt]
        \multirow{2}{*}{\textbf{Method}}  &\multicolumn{2}{c}{\textbf{Inference Latency~(ms)}}&\textbf{FLOPS}&\multirow{2}{*}{\textbf{BD-Rate}}\\
        \cmidrule(lr){2-3}
        &\textbf{Enc.}&\textbf{Dec.}& \textbf{(/G)}&   \\
        \midrule
        VVC~\cite{bross2021overview}&- &-& -&0.00  \\
        \hline
       ELIC~\cite{he2022elic} &40.76&45.34&109.38& -3.95\%\\
       Contextformer$^\dag$~\cite{koyuncu2022contextformer} &40.00& 44.00&-&-5.05\%\\
       MambaVC~\cite{qin2024mambavc} &60.45& 41.67&135.64&-7.31\% \\
       Mixed~\cite{liu2023learned} &127.36 &91.44 &211.54&-7.39\%\\
       \textbf{Ours} &60.73& 39.42& 202.04& \textbf{-12.52\%} \\ 
    \bottomrule[1.3pt]
    \end{tabular}}
    \vspace{-15pt}
    \label{tab:decoding}
\end{table}

\noindent \textbf{Inference latency.} With the increasing demand for real-time transmission, the speed of inference, especially the decoding time, greatly determines the utility of the actual scenario applications. In~\Cref{tab:decoding}, we compare the inference efficiency, \eg, encoding and decoding time of different approaches. Concretely, due to the well-designed context model with window-based local attention for entropy modeling and latent representation decoding, \modelname{} achieves higher performance, \eg, BD-rate, and lower decoding latency without sacrificing the speed of encoding.
It is worth highlighting that compared to state-of-the-art mixed transformer-CNN architectures~\cite{liu2023learned}, \modelname{} achieves notable \textbf{2.87\%} higher BD-rate with around \textbf{50\%} of encoding time and \textbf{30\%} of decoding time, firmly demonstrating the efficacy and efficiency of the proposed method.

\begin{table}[]
    \centering
    \caption{Ablation study of each component in \modelname{}. Results are evaluated on Kodak. BD-rate is compared to VVC. CAM and WLA refers to channel-wise auto-regressive modeling and window-based local attention, respectively.}
    \vspace{-5pt}
    \renewcommand{\arraystretch}{1.2}
    \resizebox{\linewidth}{!}{
    \begin{tabular}{cccc} 
    \toprule[1.3pt]
    \multirow{2}{*}{\textbf{Method}}  &\multicolumn{2}{c}{\textbf{Inference Latency~(ms)}}&\multirow{2}{*}{\textbf{BD-Rate}}\\
        \cmidrule(lr){2-3}
        &\textbf{Enc.}&\textbf{Dec.}  \\
        \midrule
        VVC~\cite{bross2021overview} &- & -& 0.00\\
        \midrule
       w/o CAM & 23.41	&16.72&-6.73\%\\
       w/o spatial context model  &46.51&32.73&-8.54 \\ 
       w/o WLA & 58.40&35.14&-9.17\% \\
       \midrule
       \textbf{Ours} & 60.73 &39.42\% &\textbf{-12.52\%}\\ 
    \bottomrule[1.3pt]
    \end{tabular}}
    \label{tab:ablation-component}
    \vspace{-5pt}
\end{table}

\begin{table}[t]
    \centering
    \caption{Effectiveness of SSM block in the model. We implement two variants that replace SSM with CNN/Transformer.}
    \vspace{-5pt}
    \renewcommand{\arraystretch}{1.2}
    \resizebox{\linewidth}{!}{
    \begin{tabular}{c cc} 
    \toprule[1.3pt]
        \multirow{2}{*}{\textbf{Foundation Blocks}}  & \textbf{Decoding}  &\multirow{2}{*}{\textbf{BD-rate}}\\
        &\textbf{Latency~(ms)}&\\
        \midrule
        CNN&\textbf{35.53}&-3.81\%\\
        Transformer &48.74&-7.19\%\\
        \textbf{State Space Model~(Ours)}&39.42&\textbf{-12.52\%} \\
    \bottomrule[1.3pt]
    \end{tabular}}
    \label{tab:ablation-foundation-block}
    \vspace{-15pt}
\end{table}

\subsection{Ablation Study}
We comprehensively analyze the influence of different structures and hyper-parameters in the proposed framework. More results can be found in~\Cref{sec:additional-results}.

\noindent \textbf{Impact of each component.} We study the effectiveness of each component proposed for enhancing image compression with SSMs and showcase the results in ~\Cref{tab:ablation-component}. It is elucidated that each designed component of our approach are of great significance to high-performance image compression. Specifically, all added modules do not bring significant delay increases and play a vital role in improving compression performance~(5.79\%, 3.98\% and 3.35\%, respectively). Moreover, compared to CAM that adds 22.70 ms latency, the proposed context model and window-based local attention only result in relatively small decoding latency~(6.69 and 4.28 ms, respectively), which can be attributed to the efficiency of compact latent representation and SSM block in the proposed framework. Finally, integrating all of them, \ie, \modelname{} achieves the best results, substantiating that the proposed context modeling techniques tailored for SSMs contribute fundamentally to the enhancement of latent representation coding and thus achieve high-performance image compression.

\noindent \textbf{Effectiveness of SSM block.} The composition of SSM block in compression model is of vital importance to enhancing the performance of image compression. To validate the utility of SSM block in boosting rate-distortion performance, we implement two variants that replace the foundation blocks of \modelname{} from SSM block to CNN and transformer, respectively. As shown in~\Cref{tab:ablation-foundation-block}, our method with SSM block achieves substantial improvements against the two variants regarding rate-distortion balance, \ie, BD-rate, showcasing the utility of the proposed techniques tailored for enhancing performance of SSM blocks. In terms of efficiency, CNN exceeds all, closely followed by SSM, with the transformer exhibiting the lowest efficiency. The outcome certifies that, considering both, the SSM block achieves a better balance between efficiency and performance and has the potential to become a foundation block in image compression.

\begin{table}[t]
    \centering
    \caption{Bitstream comparison of different methods with PSNR approximately 34.2 dB on Kodak. Previous methods are either equipped solely with local or global receptive fields. Our method benefits from local attention for performance enhancement.}
    \vspace{-5pt}
    \renewcommand{\arraystretch}{1.2}
    \resizebox{\linewidth}{!}{
    \begin{tabular}{c cccc} 
    \toprule[1.3pt]
        \multirow{2}{*}{\textbf{Method}}  & \textbf{Receptive} & \textbf{Window}  &\multirow{2}{*}{\textbf{Bpp}}&\textbf{$\Delta$Bpp}\\
        & \textbf{fields}& \textbf{size}&&(\%)\\
        \midrule
        ELIC~\cite{he2022elic} &local&-&0.4683&-\\
        Contextformer~\cite{koyuncu2022contextformer}&global&-&0.4596& 1.86\% \\
        MambaVC~\cite{qin2024mambavc} &global&-&0.4482&4.29\%\\
       \midrule
       \multirow{3}{*}{\textbf{Ours}}
        &global& 6$\times$6& 0.4428&5.44\% \\ 
        &+& 8$\times$8&\textbf{0.4404}&\textbf{5.95}\%\\
        &local&10$\times$10&0.4417&5.68\%\\
       
    \bottomrule[1.3pt]
    \end{tabular}}
    \label{tab:ablation-bitstream}
    \vspace{-15pt}
\end{table}

\begin{figure*}[t]
    \centering
    \includegraphics[width=\linewidth]{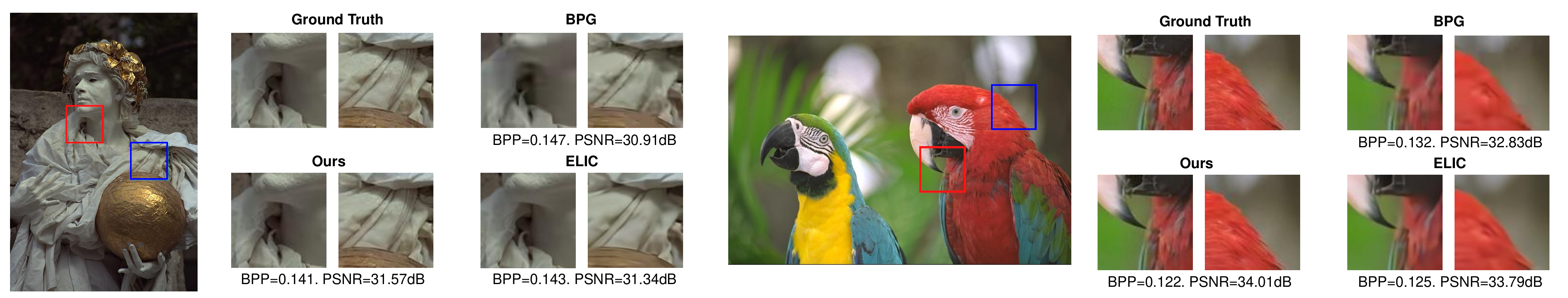}
    \vspace{-15pt}
    \caption{Visualization of decompressed images \textit{kodim17.png} and \textit{kodim23.png} in Kodak using different compression strategies. The local details are enlarged to better highlight the difference. Zoom in for a better view.}
    \vspace{-5pt}
    \label{fig:reconstructed-image}
\end{figure*}

\begin{figure*}[t]
    \centering
    \includegraphics[width=0.80\linewidth]{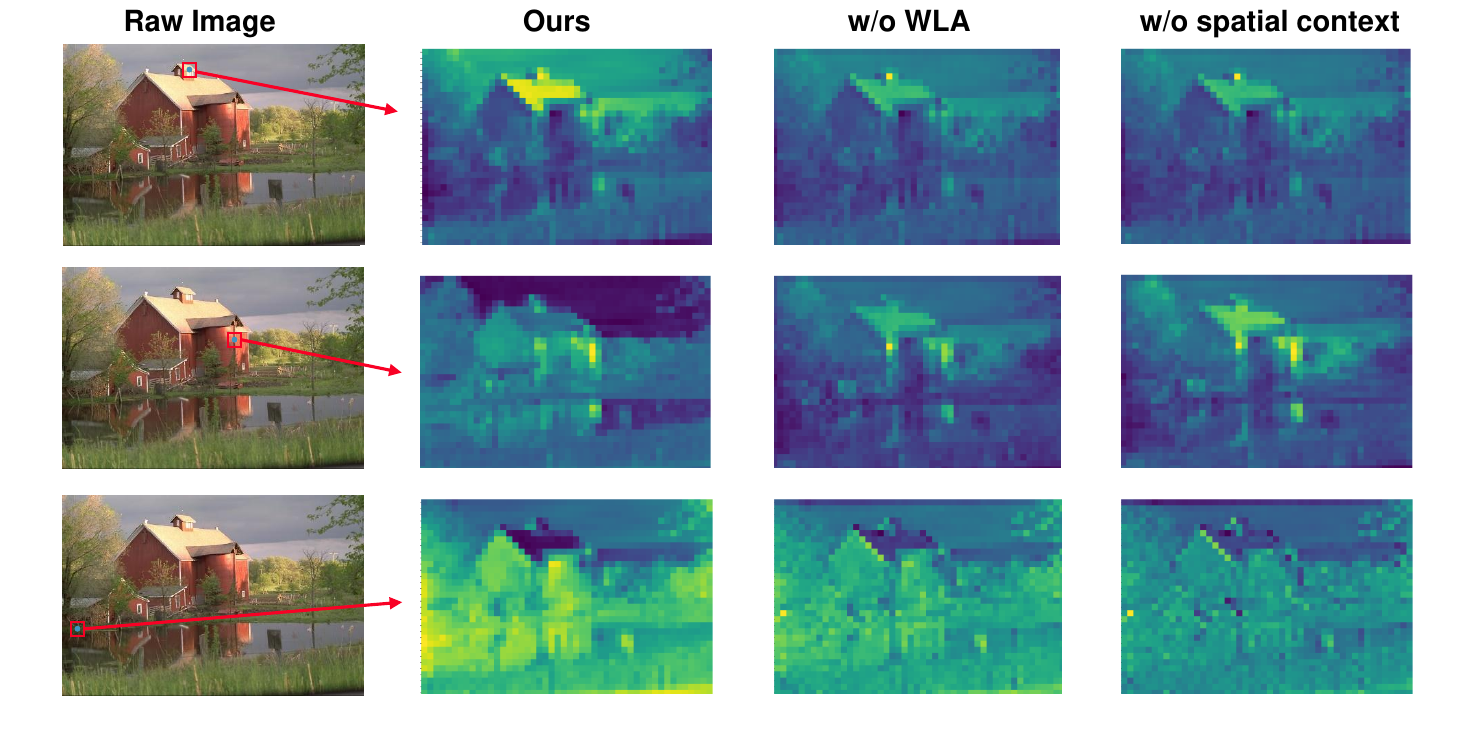}
    \vspace{-10pt}
    \caption{Attention maps of latent representations $y$ of \textit{kodim22.png} in Kodak focusing on typical points spatially. WLA represents the proposed window-based local attention. Equipped with context entropy modeling and window-based local attention, key points of \modelname{} focus on regions with strong spatial relation and therefore the method achieves compact and high-performance latent representation.}
    \vspace{-15pt}
    \label{fig:visual-attention-map}
\end{figure*}

\noindent \textbf{Influence of local window size.}
Window-based local attention is of crucial significance in mitigating the spatial redundancy and obtaining compact bitstream representation. It is shown in~\Cref{tab:ablation-bitstream} that first, previous methods possess global or local receptive fields, leading to insufficient and inefficient bitstream coding. By contrast, our approach enhances the global receptive fields of SSMs with window-based local attention, which significantly contributes to compact bitstream representation~(5.95\% bitstream saving). Moreover, window size has a moderate influence on the performance. Considering that a larger window size may not obtain information with compact local relation and a smaller window size may have a negative impact on the balance between global receptive field and local redundancy elimination, we select the window size to be 8$\times$8 in the main experiments.

\subsection{Qualitative Analysis}
\label{sec:qualitative-analysis}

In additional to quantitative experiments, we provide further qualitative results to intuitively demonstrate the effectiveness of the proposed method.

\noindent \textbf{\modelname{} achieves high quality decompression performance.} To showcase deeper analysis of the proposed \modelname{}, we visualize a few compressed images after decoding and present qualitative results in~\Cref{fig:reconstructed-image} to have a better understanding of the performance. Intuitively, compared to existing methods, \modelname{} shows obvious high-quality decompression performance with fine local features and edge details. Specifically, compared to BPG and ELIC, our image demonstrates a superior restoration effect of the sculpted chin and clothes, making them more realistic~(left of~\Cref{fig:reconstructed-image}). Additionally, it offers improved consistency and a more detailed depiction of the parrot's hair and body edges~(right of~\Cref{fig:reconstructed-image}). The outcomes strongly certificate that \modelname{} is capable of guaranteeing high-performance image compression.

\noindent \textbf{\modelname{} effectively reduces redundancy with context modeling and fine local attention.} As stated in~\Cref{sec:mambaic}, both context modeling and window-based local attention tailored for SSM-based image compression are of great significance for spatial relation description as well as high performance. We also visualize the attention map of the images decompressed by different methods and focus on several key points in~\Cref{fig:visual-attention-map}. It can be seen that no matter which points the models pay attention to, our method always better captures exact local regions, \ie, the roof~(\textit{top}), the main body of the house~(\textit{middle}), and trees and reflections in the lake~(\textit{bottom}), with similar and strong-correlated content information. For example, our method precisely and effectively enhances attention on the house~(the second row), while contrast methods focus their attention on the sky. In the third row, \modelname{} accurately identify regions with strong correlated, while contrast methods present a wide and divergent attention to the entire image. It is illustrated that, unlike an approach without window attention, \modelname{} better helps focus on crucial spatial points and therefore eliminate redundancy as much as possible. Moreover, context modeling contributes to latent representation with rich side information, thereby enhancing compact local relations. All visualizations vividly certificate the utility of the proposed method for compact latent representation and therefore achieving high-performance image compression.

\section{Conclusion}
In this paper, we explore SSM for high-performance learned image compression and introduce \modelname{} to tackle the problem of \textit{computational efficiency} and \textit{effectiveness}. In particular, we introduce context modeling into SSM-based compression pipeline, which empowers entropy modeling with rich side information. Then an efficient window-based local attention is employed to eliminate spatial redundancy. Both qualitative and quantitative experimental results certificate the effectiveness against existing methods with superior efficiency, and the improvement is especially substantial with higher resolution.

\section*{Acknowledgements}
This work is supported by Wuxi Research Institute of Applied Technologies, Tsinghua University under Grant 20242001120, and the Fundamental Research Funds for the Central Universities, Peking University.

{
    \small
    \bibliographystyle{ieeenat_fullname}
    \bibliography{main}
}

\clearpage
\setcounter{page}{1}
\setcounter{figure}{0}
\setcounter{table}{0}
\renewcommand{\thefigure}{A\arabic{figure}}
\renewcommand{\thetable}{A\arabic{table}}
\maketitlesupplementary
\appendix

In the appendix, we provide details about evaluation metrics~(\cref{sec:appendix-metrics}), datasets~(\cref{sec:appendix-datasets}) and experimental settings~(\cref{sec:appendix-settings}). We also carry out more experimental results~(\cref{sec:additional-results}) and visualizations~(\cref{sec:appendix-heatmaps}) to showcase the effectiveness of the proposed method qualitatively and quantitatively.

\section{Explanation of Evaluation Metrics}
\label{sec:appendix-metrics}
\noindent \textbf{PSNR.} Peak Signal-to-Noise Ratio~(PSNR) is a widely used metric to measure the quality of reconstructed images compared to the original images. It quantifies how much the noise~(\ie, distortion) has affected the quality of the image. Higher PSNR values typically indicate better quality, with less distortion or degradation in the image. In the main results, we convert PSNR into a logarithmic decibel unit for a better comparison.

\noindent \textbf{MS-SSIM.} Multiscale Structural Similarity Index~(MS-SSIM) is an extension of SSIM~(Structural Similarity Index). Concretely, SSIM evaluates the perceived quality of an image based on three main factors: luminance, contrast, and structure. The combination of these three components gives a measure of image quality that aligns more closely with human perception. Furthermore, MS-SSIM improves upon the original metric by evaluating similarity at multiple scales~(resolutions) to better simulate human perception. In practical calculations, MS-SSIM combines the SSIM values from different scales using a weighted average.

\section{Details about Evaluation datasets}
\label{sec:appendix-datasets}
\noindent \textbf{Kodak.} kodak is made up of 24 high-quality color images, each of them with 768 × 512 pixels. These images contains a diverse set of scenes, including landscapes, portraits, indoor settings, and textures, making the dataset representative of real-world visual content.

\noindent \textbf{Tecnick.} Tecnick consists of 100 images with  $1200\times1200$ pixels. It is significant in evaluating image compression performance on numerous images with medium resolution.

\noindent \textbf{CLIC Professional Valid.} CLIC Professional Valid is a collection of images with 2K resolution proposed by the Third Challenge on Learned Image Compression. It validates the effectiveness of learned image compression approaches on high-resolution scenarios.

\section{More Explanation of Experimental Settings}
\label{sec:appendix-settings}
Detailed structure of channel-spatial context model is shown in~\Cref{tab:appendix-architecture}. In the main paper, structure of hyper encoder/decoder are a stack of convolution/deconvolution, VSS block and convolution/deconvolution. The convolution in spatial and channel entropy modeling $\Phi$ and $\Psi_k$ holds \texttt{kernel=3, stride=2} by default. In training procedure, the images are randomly cropped to $256\times256$. We use Adam optimizer with $\beta_1=0.9$, $\beta_2=0.999$. The learning rate is set to \text{1e-4} by default. During evaluation, the image is padded to fit for the compression and all evaluations are conducted on NVIDIA A100 under the same condition.

\begin{table}[h]
    \centering
    \caption{Detailed architecture of channel-spatial context model.}
    \vspace{-5pt}
    \footnotesize
    \resizebox{\linewidth}{!}{
    \begin{tabular}{c|c}
    \toprule[1.3pt]
         \textbf{Spatial context model $\Phi$}& \textbf{Channel context model $\Psi_k$}\\
         \midrule
         \makecell{in channel: M/K \\
         (spatial, $K=5$)} & \makecell{in channel: k*M/K\\($k^{th}$, channel $k=1,\cdots,K$)}\\ 
        \midrule
        \makecell{VSS block \\ Conv $3\times3$, s1, 2*M/K}&\makecell{ VSS block \\ Conv $3\times3$, s1, 2*M/K}\\
        \midrule
        \midrule
        \multicolumn{2}{c}{\textbf{WLA module with channel-spatial aggregation}} \\
        \midrule
        \multicolumn{2}{c}{\makecell{fixed channel: 2*M/K+2*M/K+2*M
         \\ (spatial+channel+hyper) spatial reshape} } \\
         \midrule
        \multicolumn{2}{c}{\makecell{partition window size $w$ \\ Local Attention \\ reverse window size $w$}} \\
        \bottomrule[1.3pt]
    \end{tabular}}
    \label{tab:appendix-architecture}
    \vspace{-15pt}
\end{table}

\section{More Experimental Results}
\label{sec:additional-results}
\noindent \textbf{Effectiveness of SSM block in different modules.} We apply SSM block as foundation block in both nonlinear transform and context model. To figure out the utility of different foundation blocks in each modules, we additional conduct experiments and comprehensively compare the results with different variants of model that is equipped with nonlinear transform/context model of CNN/Transformer/SMM structure. Results in~\Cref{tab:appendix-context-model} reveals that the structure of main transform, \ie, encoder/decoder, also influences the performance and further demonstrates the effectiveness of the proposed structure incorporating SSM blocks.

\section{More Attention Map Comparison}
\label{sec:appendix-heatmaps}
\begin{figure*}[ht]
    \centering
    \includegraphics[width=0.8\linewidth]{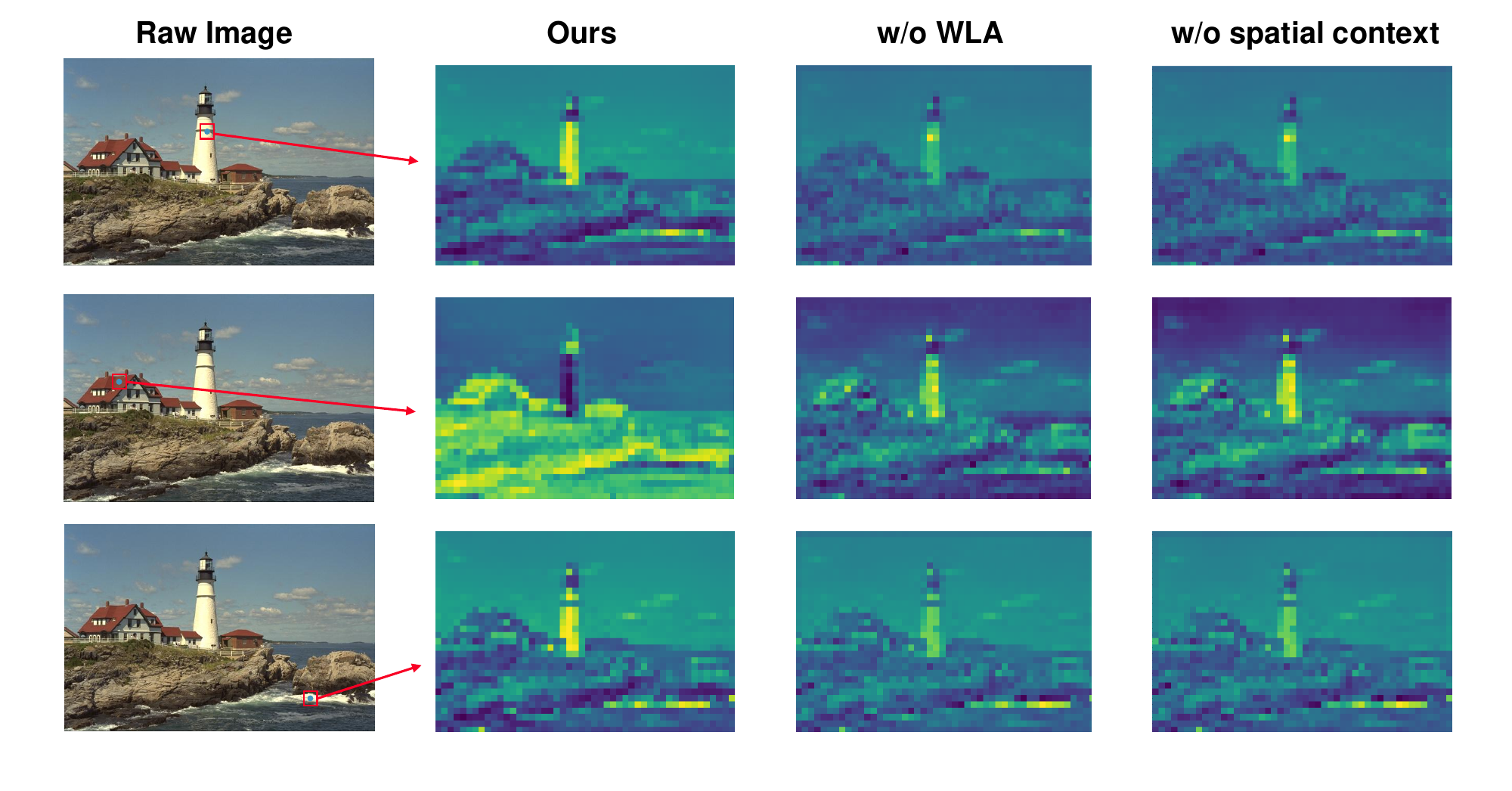}
    \vspace{-15pt}
    \caption{Attention maps of latent representations $y$ of \textit{kodim21.png} in Kodak.}
    \label{fig:appendix-attn-1}
\end{figure*}

\begin{figure*}[ht]
    \centering
    \includegraphics[width=0.8\linewidth]{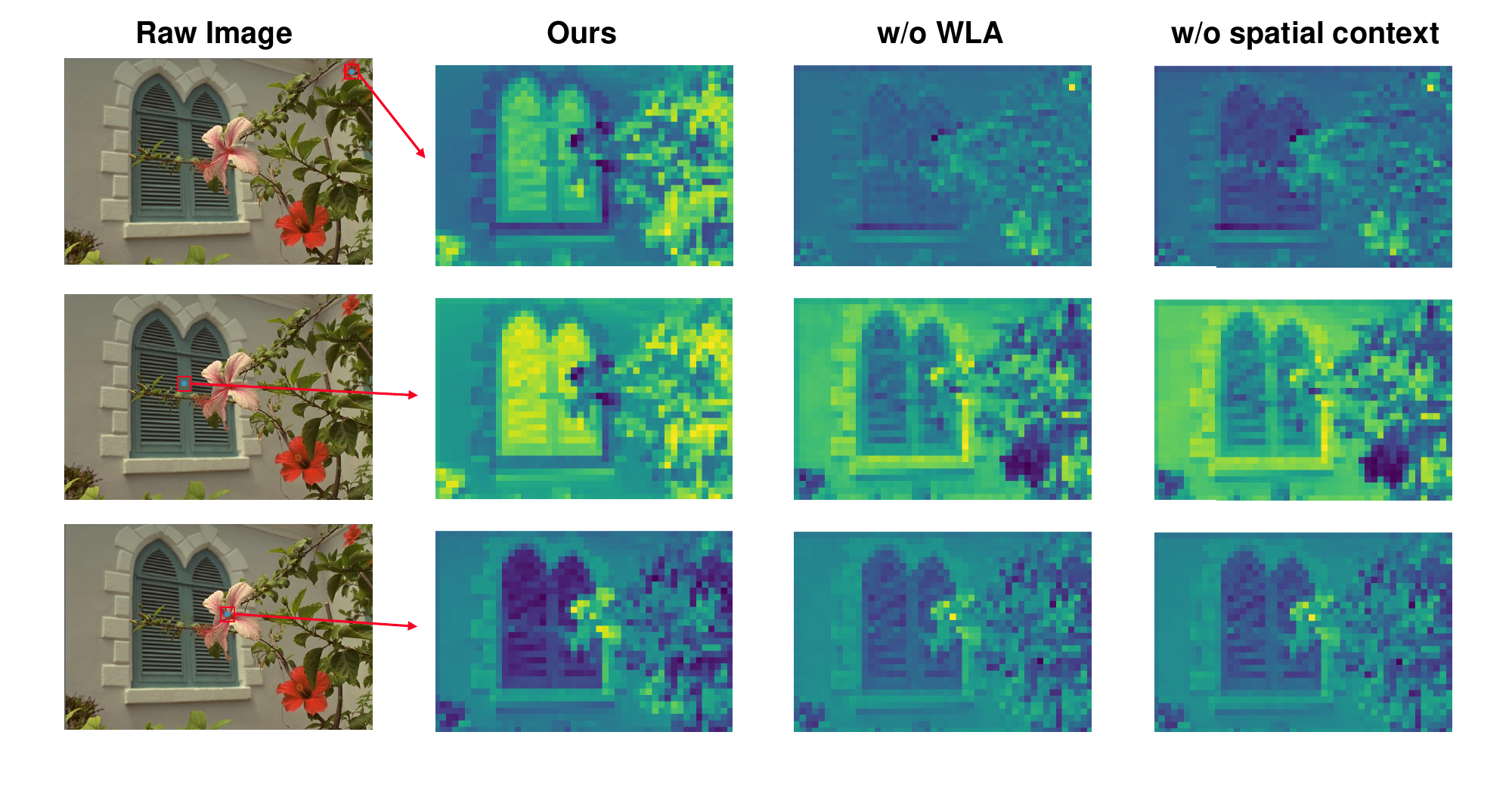}
    \vspace{-15pt}
    \caption{Attention maps of latent representations $y$ of \textit{kodim07.png} in Kodak.}
    \label{fig:appendix-attn-2}
\end{figure*}

Corresponding to visualization results in the main paper, we showcase more comparison of attention map as opposed to models w/o context entropy model and w/o window-based local attention in~\Cref{fig:appendix-attn-1} and~\Cref{fig:appendix-attn-2} to further verify the effectiveness of each proposed component in \modelname{}.

\begin{table*}[h]
    \centering
    \caption{Different variants of nonlinear transform architecture.}
    \vspace{-5pt}
    \setlength{\tabcolsep}{8pt}
    \renewcommand{\arraystretch}{1.25}
    \resizebox{0.9\linewidth}{!}{
    \begin{tabular}{c cccc}
    \toprule[1.3pt]
        \textbf{Main Transform} & \textbf{Hyper Transform}& \textbf{Context Model}&\textbf{Decoding Latency~(ms)} &\textbf{BD-Rate}\\
        \hline
        \multirow{3}{*}{CNN}&\multicolumn{2}{c}{CNN}& 35.53&-3.81\%\\
        &\multicolumn{2}{c}{Transformer}&-&-\\
        &\multicolumn{2}{c}{State Space Model}&35.64 & -7.15\%\\
        \hline
         \multirow{3}{*}{Transformer}&\multicolumn{2}{c}{CNN}&-&-\\
         & \multicolumn{2}{c}{Transformer}&48.74&-7.19\%\\
        &\multicolumn{2}{c}{State Space Model}&37.82 & -9.30\%\\
          \hline
          \multirow{3}{*}{State Space Model}&\multicolumn{2}{c}{CNN}&-&- \\
          &\multicolumn{2}{c}{Transformer}&-&-\\
          &\multicolumn{2}{c}{State Space Model}&39.42&-12.52\%\\
    \bottomrule[1.3pt]
    \end{tabular}}
    \vspace{-10pt}
    \label{tab:appendix-context-model}
\end{table*}

\end{document}